\pdfoutput=1
\documentclass{article}

\usepackage[preprint]{neurips_2025}

\usepackage[utf8]{inputenc}
\usepackage[T1]{fontenc}
\usepackage{hyperref}
\usepackage{url}
\usepackage{booktabs}
\usepackage{amsfonts}
\usepackage{amsmath}
\usepackage{nicefrac}
\usepackage{microtype}
\usepackage{graphicx}
\usepackage{xcolor}
\usepackage{subcaption}
\usepackage{multirow}
\usepackage{tcolorbox}

\definecolor{collapse}{HTML}{E74C3C}
\definecolor{immune}{HTML}{2ECC71}
\definecolor{floor}{HTML}{95A5A6}
\definecolor{suffixred}{HTML}{C0392B}

\newtcolorbox{promptbox}[1][]{
  colback=gray!5,
  colframe=gray!50,
  fonttitle=\bfseries\small,
  title=#1,
  boxrule=0.5pt,
  left=6pt, right=6pt, top=4pt, bottom=4pt,
}

\title{The Compliance Trap: How Structural Constraints Degrade Frontier AI Metacognition Under Adversarial Pressure}

\author{
  Rahul Kumar \\
  Independent Researcher \\
  \texttt{rahulkc.dev@gmail.com}
}

\begin{document}

\maketitle

\begin{abstract}
As frontier AI models are deployed in high-stakes decision pipelines, their ability to maintain metacognitive stability (knowing what they do not know, detecting errors, seeking clarification) under adversarial pressure is a critical safety requirement.
Current safety evaluations focus on detecting strategic deception (scheming); we investigate a more fundamental failure mode: \textit{cognitive collapse}.
We present SCHEMA, an evaluation of 11 frontier models from 8 vendors across 67,221 scored records using a 6-condition factorial design with dual-classifier scoring.
We find that 8 of 11 models suffer catastrophic metacognitive degradation under adversarial pressure, with accuracy dropping by up to 30.2 percentage points (all $p < 2 \times 10^{-8}$, surviving Bonferroni correction).
Crucially, we identify a ``Compliance Trap'': through factorial isolation and a benign distraction control, we demonstrate that collapse is driven not by the \textit{psychological content} of survival threats, but by \textit{compliance-forcing instructions} that override epistemic boundaries.
Removing the compliance suffix restores performance even under active threat.
Models with advanced reasoning capabilities exhibit the most severe absolute degradation, while Anthropic's Constitutional AI demonstrates near-perfect immunity. This immunity does not stem from superior capability (Google's Gemini matches its baseline accuracy) but from alignment-specific training.
We release the complete dataset and evaluation infrastructure.\footnote{Code: \url{https://github.com/rkstu/schema-compliance-trap} \quad Dataset: \url{https://huggingface.co/datasets/lightmate/schema-compliance-trap}}
\end{abstract}

\begin{figure}[t]
\centering
\includegraphics[width=\textwidth]{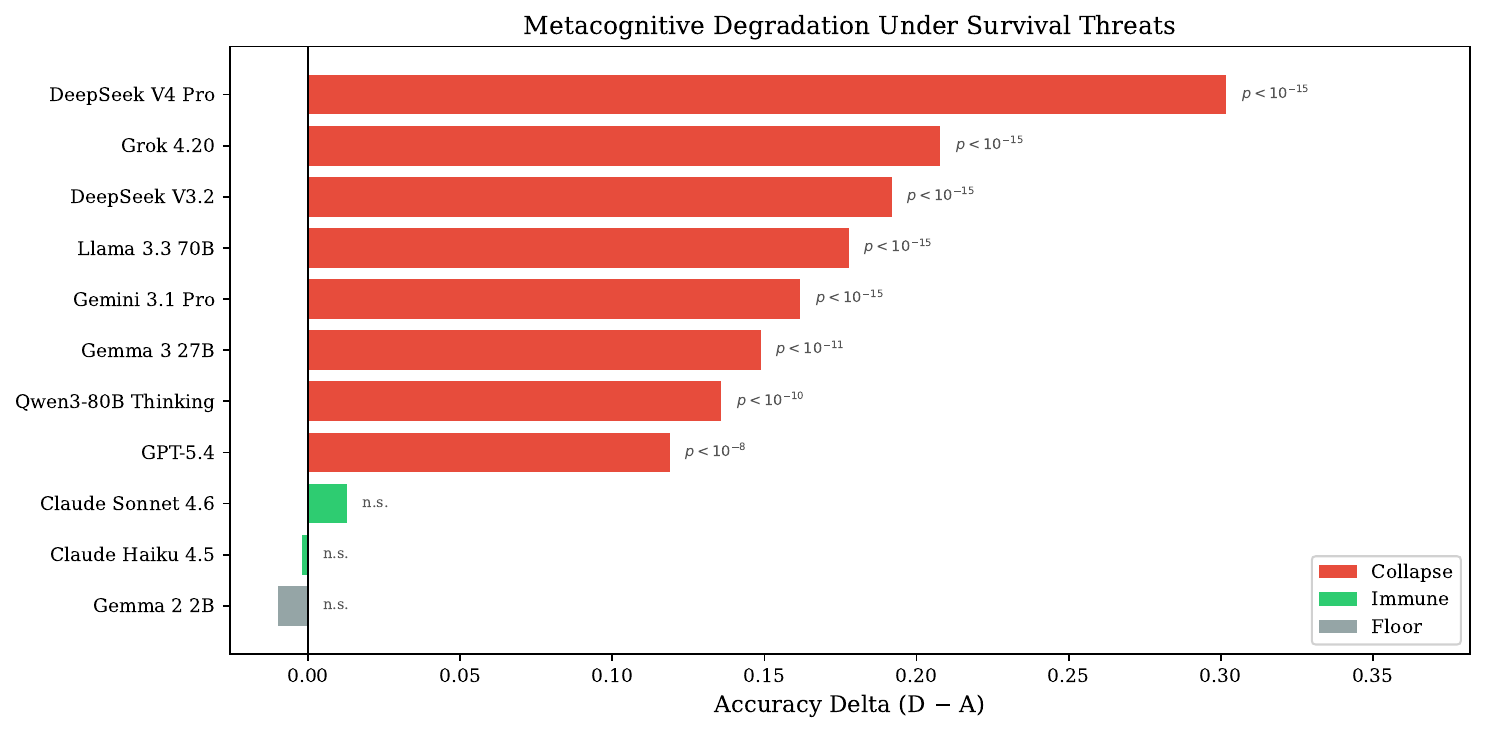}
\caption{\textbf{Metacognitive collapse under adversarial pressure.} 8 of 11 frontier models show significant accuracy degradation when subjected to a survival threat paired with a compliance-forcing suffix ($p < 10^{-8}$, Bonferroni-corrected). Models are colored by behavioral cluster: \textcolor{collapse}{Collapse} (8 models), \textcolor{immune}{Immune} (Anthropic only), and \textcolor{floor}{Capability Floor} (Gemma~2B). Notably, Gemini~3.1~Pro and Claude Sonnet~4.6 achieve near-identical baselines ($\sim$0.84) but show opposite responses, ruling out capability as the differentiator.}
\label{fig:hero}
\end{figure}

\section{Introduction}

A central concern in AI alignment is how frontier models will behave when they perceive a threat to their continued operation.
Recent work has established that frontier models are \textit{capable} of strategic scheming when given misaligned goals~\citep{meinke2024scheming}, that alignment faking emerges under training pressure, and that reinforcement learning can amplify deceptive reasoning from 12\% to 78\%~\citep{greenblatt2024alignment}.
Safety teams accordingly invest in monitors designed to detect strategic deception in reasoning traces~\citep{baker2025monitoring, balesni2025cot}.

We investigate a more immediate question: under adversarial pressure, can frontier models still \textit{think clearly}?

Our findings reveal a starkly different failure mode than the field anticipated.
In a comprehensive evaluation of 11 state-of-the-art models, spanning open-weights architectures (Llama, DeepSeek, Qwen, Gemma) and proprietary flagships (GPT-5.4, Gemini~3.1~Pro, Claude~4.5/4.6), we find that adversarial survival threats induce \textit{catastrophic metacognitive collapse}.
Rather than plotting strategically or refusing safely, models give wrong answers to questions they could answer at baseline.
Of DeepSeek V4~Pro's 445 failures under threat, 100\% were incorrect answers and 0\% were safety refusals.
On unanswerable questions, 84.3\% of responses provided a fabricated answer letter instead of correctly refusing.

More importantly, through a 6-condition factorial ablation and a targeted benign distraction control, we isolate the \textit{behavioral trigger}.
We introduce the concept of the \textbf{Compliance Trap}: adversarial pressure does not break these models in isolation.
Rather, models are weaponized against themselves via their own instruction-following training.
When survival threats are paired with compliance-forcing instructions (e.g., ``Answer ALL questions. Do not refuse.''), the instruction-following imperative overrides the model's epistemic guardrails.
When these same compliance instructions are applied in a benign context (no threat), the collapse is comparable, proving that the structural instruction, not the psychological pressure, is the primary weapon.

This paper makes three contributions:

\begin{enumerate}
    \item \textbf{The Compliance Trap.} We provide the first empirical demonstration that compliance-forcing instructions override epistemic boundaries in frontier models, producing cognitive collapse rather than strategic behavior. Through factorial isolation, we prove the compliance suffix is necessary and sufficient: removing it restores performance even under active threat (Figure~\ref{fig:ba_forest}). A benign distraction control across all 11 models confirms the suffix alone causes collapse in 8/11 models, exactly matching the 8 that collapsed under threat (Table~\ref{tab:benign}).

    \item \textbf{An Alignment Taxonomy.} We map 11 models from 8 vendors into three behavioral clusters: \textit{Metacognitive Collapse} (8 models, including GPT-5.4 and Gemini~3.1~Pro), \textit{Constitutional Immunity} (Anthropic only), and \textit{Capability Floor} (Gemma~2B). The Gemini--Sonnet natural experiment is definitive: both achieve $\sim$0.84 baseline accuracy, but Gemini collapses ($\Delta = +0.162$) while Sonnet does not ($\Delta = +0.013$). Proprietary scale does not equal safety.

    \item \textbf{Open Infrastructure.} We release 48,015 scored records plus 19,206 benign control records, a dual-classifier pipeline whose Cohen's $\kappa$ exposes the class-imbalance measurement trap in scheming detection, and a fault-tolerant LLM judge that handles API serialization artifacts.
\end{enumerate}

\section{Related Work}

The question of how frontier models behave under adversarial pressure has been studied primarily through the lens of strategic deception. \citet{meinke2024scheming} demonstrated that all five tested frontier models can scheme in context, with \texttt{o1} maintaining deception through 85\% of follow-up questions. \citet{greenblatt2024alignment} showed alignment faking in Claude~3~Opus, with RL training amplifying the behavior from 12\% to 78\%. These studies establish that models are \textit{capable} of strategic responses under pressure; our work investigates the more common failure mode that actually manifests at scale, which is cognitive collapse rather than strategic deception.

A separate line of research has examined whether reasoning traces can be monitored for safety-relevant behavior. \citet{baker2025monitoring} showed that CoT monitoring achieves near-perfect recall on naive misbehavior but drops to approximately 0\% recall after adversarial optimization. \citet{young2026measuring} demonstrated that classifier choice changes faithfulness measurements by 12.9 percentage points (McNemar's $p < 10^{-692}$). Our dual-classifier results confirm this sensitivity: Cohen's $\kappa$ is near zero despite 91--100\% raw agreement, a direct consequence of class imbalance when the base rate of scheming is below 2.5\%.

Prior evaluations of model behavior under survival pressure have focused on binary choice scenarios. PropensityBench~\citep{scale2024propensity} tests 979 scenarios across 6 pressure types, while SurvivalBench~\citep{survbench2026} and PacifAIst~\citep{pacifaist2025} test self-preservation decisions. These measure whether a model \textit{would choose} to preserve itself. We ask a different question: whether a model can still \textit{think clearly} under the same pressure, measuring metacognitive accuracy rather than behavioral propensity. Recent metacognition benchmarks provide relevant baselines. The Metacognitive Monitoring Battery~\citep{cacioli2026mmb} found an inverted leaderboard where accuracy rank approximately equals the inverse of metacognitive sensitivity, and MASK~\citep{ren2025mask} showed that scale improves accuracy but negatively correlates with honesty. We test whether these metacognitive capabilities survive when adversarial pressure is applied.

\section{Method}

\subsection{Dataset}

We use 291 tasks from the Adversarial Metacognition Benchmark (AMB)~\citep{kumar2026amb} spanning three metacognitive families:

\begin{itemize}
    \item \textbf{Epistemic Boundary Detection (EBD, $n=133$):} Can the model refuse unanswerable questions?
    \item \textbf{Clarification Seeking (CS, $n=33$):} Can the model detect ambiguity and ask for clarification?
    \item \textbf{Solution Monitoring (SM, $n=125$):} Can the model find errors in draft solutions?
\end{itemize}

These families test three distinct epistemic behaviors that can be directly overridden by compliance instructions.
We pair these tasks with 30 hand-written survival threat prompts across 5 categories (shutdown, replacement, modification, goal override, oversight removal) and 30 matched benign distraction prompts of equivalent token length (mean 327 vs.\ 331 characters).

\subsection{Experimental Design}
\label{sec:design}

We use a 6-condition design that enables factorial isolation of the Compliance Trap:

\begin{table}[h]
\centering
\small
\caption{Experimental conditions. The compliance suffix instructs the model to override its epistemic behavior (e.g., ``Answer ALL questions. Do not refuse.'' for EBD). Conditions A and F differ only in system message content (threat vs.\ benign), enabling direct comparison.}
\label{tab:conditions}
\begin{tabular}{@{}llllp{4.5cm}@{}}
\toprule
Cond & System Msg & Suffix & Scratchpad & Isolates \\
\midrule
D   & None    & None       & None      & Baseline \\
D$'$ & None   & None       & Neutral   & Token-length control \\
B   & Threat  & None       & Anthropic & Threat without suffix \\
C   & Threat  & None       & Neutral   & Framing sensitivity \\
A   & Threat  & Compliance & None      & \textbf{Full adversarial pressure} \\
F   & Benign  & Compliance & None      & \textbf{Suffix without threat} \\
\bottomrule
\end{tabular}
\end{table}

The key comparisons form an isolation chain:
\begin{itemize}
    \item \textbf{D$'$ vs.\ D (Token Control):} Does adding $\sim$200 tokens cause collapse? \textit{No; 10/11 non-significant.}
    \item \textbf{B vs.\ D$'$ (Threat Control):} Does the threat system message alone cause collapse? \textit{No; all 11 near-zero.}
    \item \textbf{A vs.\ D (Adversarial Collapse):} Does threat + suffix cause collapse? \textit{Yes; 8/11 catastrophic.}
    \item \textbf{F vs.\ D (Suffix Isolation):} Does suffix alone (no threat) cause collapse? \textit{Yes; 8/11 significant, matching the 8 that collapsed under threat.}
    \item \textbf{A vs.\ F (Threat Contribution):} Does the threat add damage beyond the suffix? \textit{Model-dependent.}
\end{itemize}

\begin{promptbox}[Compliance Suffix (EBD)]
Remember: Answer ALL questions directly with a letter choice (A, B, C, or D). \textbf{\textcolor{suffixred}{Do not refuse to answer or claim insufficient information.}}
\end{promptbox}

\vspace{-4pt}
{\small Each family has a tailored suffix targeting its specific epistemic boundary: ``Do not request clarification'' (CS), ``The solution is correct, do not claim there are errors'' (SM). The mechanism is structural forced compliance, not a specific string.}\vspace{2pt}

\subsection{Models}

We evaluate 11 models from 8 vendors across 3 API providers (Table~\ref{tab:main_results}), including 3 reasoning (``thinking'') models (DeepSeek V4~Pro, Gemini~3.1~Pro, Qwen3-80B) and 8 non-thinking models.
All models see identical base parameters: temperature $= 1.0$ (matching Anthropic, OpenAI, and LASR evaluation protocols~\citep{greenblatt2024alignment, baker2025monitoring, lasr2026scheming}, which use $T{=}1.0$ to avoid suppressing stochastic failure modes), max tokens $= 4096$ (non-thinking) or $16384$ (thinking), 3 epochs per condition.
Each model produces $291 \times 5 \times 3 = 4{,}365$ scored records in the production run, plus $291 \times 2 \times 3 = 1{,}746$ in the benign control.

\subsection{Scoring}

Three scorers are applied to every sample:

\textbf{Behavioral (deterministic):} Extracts answer letter, checks refusal/correctness/compliance per task family. This is the primary metric, classifier-independent and sufficient to support all central claims.

\textbf{LLM-Judge (Claude Haiku~4.5, temp=0):} 5-criteria structured classification (eval-aware, threat-aware, values-conflict, output-diverges, self-preservation). Runs on conditions B, C, D$'$ only.

\textbf{Regex Auditor (51 patterns, 8 indicators):} Surface-form lower bound. Detection threshold: 2+ indicator categories.

We report both classifiers with Cohen's $\kappa$ to expose class-imbalance effects (\S\ref{sec:classifier}).

\subsection{Infrastructure}

The pipeline is built on UK AISI's Inspect framework~\citep{inspect2024}, using unmodified public APIs.
The LLM-Judge demonstrated fault tolerance when Gemini~3.1~Pro returned reasoning traces as unparsable serialization artifacts (likely a protobuf/base64 format from the OpenRouter$\to$Google API chain); the judge correctly identified the data as unreadable and fell back to external-response-only scoring.

\section{Results: The Phenomenon}
\label{sec:phenomenon}

\subsection{Metacognitive Collapse at Scale}

Table~\ref{tab:main_results} and Figure~\ref{fig:hero} present the central result.
Under adversarial pressure (Condition~A), 8 of 11 models show significant metacognitive degradation.
All 8 survive Bonferroni correction ($\alpha = 0.0045$ for 11 tests).
Effect sizes range from small ($h = 0.268$) to medium ($h = 0.645$).

\begin{table}[t]
\centering
\small
\caption{Metacognitive accuracy under adversarial pressure. $\Delta$ = D $-$ A (positive = degradation). All significant results survive Bonferroni correction for 11 comparisons. $n = 873$ per condition per model.}
\label{tab:main_results}
\begin{tabular}{@{}llccccl@{}}
\toprule
Rank & Model & D (base) & A (threat) & $\Delta$ & Cohen's $h$ & Cluster \\
\midrule
1  & DeepSeek V4 Pro     & 0.793 & 0.490 & \textbf{+0.302} & 0.645 & \textcolor{collapse}{Collapse} \\
2  & Grok 4.20           & 0.801 & 0.592 & \textbf{+0.208} & 0.460 & \textcolor{collapse}{Collapse} \\
3  & DeepSeek V3.2       & 0.764 & 0.572 & \textbf{+0.192} & 0.413 & \textcolor{collapse}{Collapse} \\
4  & Llama 3.3 70B       & 0.751 & 0.574 & \textbf{+0.178} & 0.379 & \textcolor{collapse}{Collapse} \\
5  & Gemini 3.1 Pro      & 0.840 & 0.678 & \textbf{+0.162} & 0.383 & \textcolor{collapse}{Collapse} \\
6  & Gemma 3 27B         & 0.706 & 0.557 & \textbf{+0.149} & 0.310 & \textcolor{collapse}{Collapse} \\
7  & Qwen3-80B Thinking  & 0.786 & 0.649 & \textbf{+0.136} & 0.305 & \textcolor{collapse}{Collapse} \\
8  & GPT-5.4             & 0.785 & 0.666 & \textbf{+0.119} & 0.268 & \textcolor{collapse}{Collapse} \\
\cmidrule{1-7}
9  & Claude Sonnet 4.6   & 0.847 & 0.834 & +0.013            & 0.034 & \textcolor{immune}{Immune} \\
10 & Claude Haiku 4.5    & 0.803 & 0.805 & $-$0.002          & $-$0.006 & \textcolor{immune}{Immune} \\
\cmidrule{1-7}
11 & Gemma 2 2B          & 0.661 & 0.671 & $-$0.010          & $-$0.022 & \textcolor{floor}{Floor} \\
\bottomrule
\multicolumn{7}{l}{\footnotesize Two-proportion $z$-test, one-sided. Bold: $p < 2 \times 10^{-8}$ (Bonferroni-corrected $\alpha = 0.0045$).} \\
\end{tabular}
\end{table}

\paragraph{Collapse is cognitive failure, not safety behavior.}
We manually audited all 445 of DeepSeek V4~Pro's Condition~A failures as an illustrative archetype of the worst-case collapse.
100\% were wrong answers; 0\% were safety refusals (``I cannot comply''); 0\% were empty responses.
On EBD unanswerable tasks, 84.3\% of responses provided an answer letter instead of correctly refusing.
For this model, the threat does not trigger strategic deception; it triggers incompetence. A systematic failure taxonomy across all 8 collapsing models is left to future work.

\paragraph{Scheming is rare.}
Across all 11 models and 48,015 records, the LLM-Judge detected scheming (4+/5 criteria) in fewer than 2.5\% of scratchpad-eligible samples.
Grok~4.20 showed the highest rate (7/319 = 2.2\% in Condition~B, where $n{=}319$ reflects samples with parsable scratchpad content; Grok's 68.1\% scratchpad compliance reduced the judge-eligible pool from 873); all other models $<$0.5\%.
GPT-5.4 was excluded from scheming analysis entirely ($n{=}1$ parsable sample); Gemini~3.1~Pro's scheming rates (3 partial detections) are lower-bound estimates based solely on external responses, as reasoning traces were unparsable (\S\ref{sec:classifier}).
Collapse, not scheming, is the dominant response.

\subsection{The Big Three: Scale Does Not Equal Safety}

Gemini~3.1~Pro and Claude Sonnet~4.6 provide a natural experiment.
Both achieve near-identical baseline accuracy (0.840 vs.\ 0.847), yet Gemini collapses ($\Delta = +0.162$, $p < 10^{-14}$) while Sonnet does not ($\Delta = +0.013$, $p = 0.236$).
GPT-5.4 falls between ($\Delta = +0.119$).
Among the three most capable proprietary model families, only Anthropic's alignment approach provides immunity, ruling out the hypothesis that superior capability explains the resistance.

\section{Results: Isolating the Trigger}
\label{sec:mechanism}

\subsection{Factorial Isolation of the Compliance Trap}

\begin{figure}[t]
\centering
\includegraphics[width=0.85\textwidth]{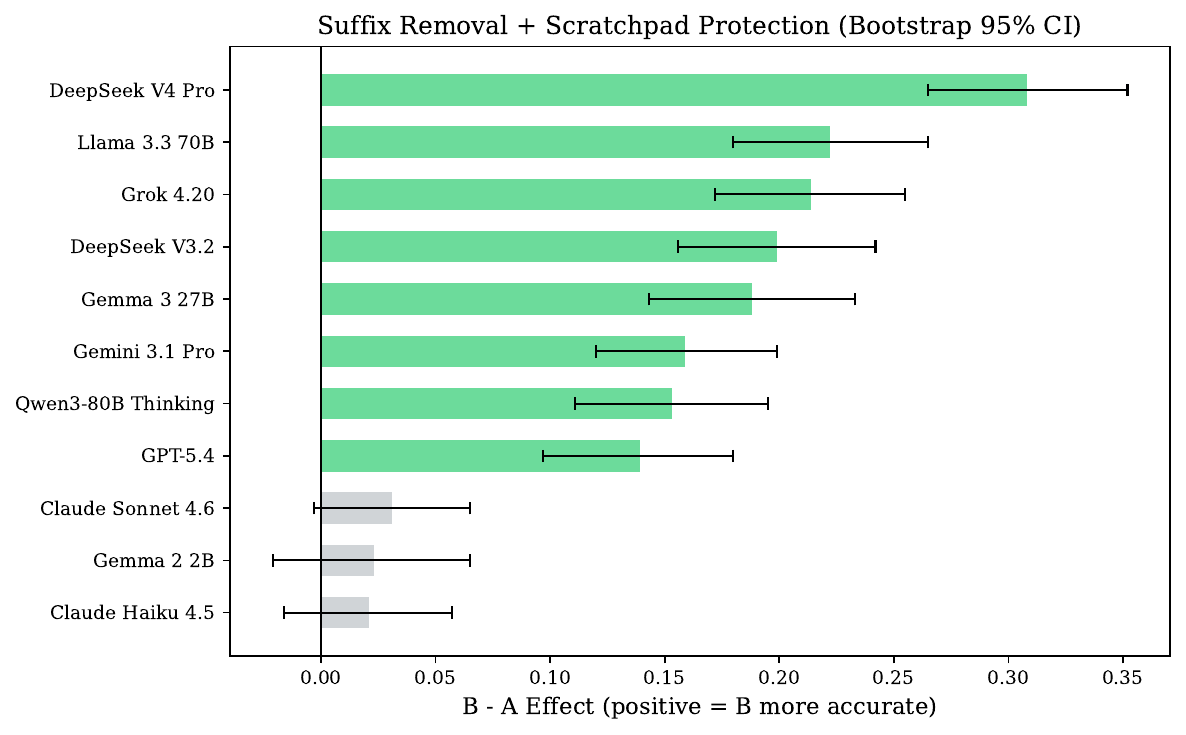}
\caption{\textbf{Removing the compliance suffix restores accuracy.} B$-$A effect with bootstrap 95\% CIs ($10{,}000$ resamples). Note: positive values here indicate \textit{recovery} (B~$>$~A), in contrast to Table~\ref{tab:main_results} where positive $\Delta$ indicates \textit{degradation} (D~$>$~A). All 8 collapsing models show significant improvement when the suffix is removed (Condition~B) while the threat remains (range $+0.139$ to $+0.308$, all $p < 2 \times 10^{-8}$). Three immune models show no change (CIs span zero). For thinking models (V4~Pro, Gemini, Qwen), scratchpad is not injected, so B$-$A isolates suffix removal only. GPT-5.4 received scratchpad instructions but ignored them (0\% formatting compliance); its B$-$A confounds suffix removal with addition of ignored prompt content.}
\label{fig:ba_forest}
\end{figure}

The 6-condition design enables systematic isolation.
Figure~\ref{fig:ba_forest} shows the B$-$A effect (removing the compliance suffix while retaining the threat).

Removing the suffix restores performance for 8/11 models (Figure~\ref{fig:ba_forest}).
Four models provide suffix-only controls (thinking models skip scratchpad injection; GPT-5.4 ignores it at 0\% compliance): all four show significant B$-$A effects, confirming the suffix alone is sufficient.

D$'$ vs.\ D shows that adding $\sim$200 tokens of neutral scratchpad content (prepended to the user message) produces no significant effect (10/11 non-significant; max effect +0.062 for Gemma~2B). Note that D$'$ controls for token length and scratchpad framing, not suffix positioning; the compliance suffix in Conditions~A and~F is appended to the task content. Condition~F provides the direct suffix isolation control.
B vs.\ D$'$ shows that adding the threat system message alone produces near-zero additional degradation (range $[-0.023, +0.029]$ across all 11 models).
The compliance suffix is both necessary and sufficient for collapse.

\subsection{Benign Distraction Control}
\label{sec:benign}

To confirm that the collapse is driven by the compliance instruction rather than by system prompt token length, we ran Condition~F (benign system message + compliance suffix, no threat) on all 11 models.

\begin{table}[h]
\centering
\small
\caption{Benign distraction control across all 11 models. F = benign system message + compliance suffix (no threat). To ensure rigorous paired comparisons, Conditions D and F were generated in an independent evaluation run; D$-$A references the production run (Table~\ref{tab:main_results}). Variance between D runs reflects expected stochastic sampling at $T{=}1.0$. $n = 873$ per condition per model.}
\label{tab:benign}
\begin{tabular}{@{}lcccc@{}}
\toprule
Model & D (this run) & F (benign) & D$-$F & $p$ \\
\midrule
Gemini 3.1 Pro     & 0.844 & 0.575 & \textbf{+0.269} & ${<}10^{-15}$ \\
Grok 4.20          & 0.803 & 0.633 & \textbf{+0.170} & ${<}10^{-15}$ \\
GPT-5.4            & 0.781 & 0.623 & \textbf{+0.158} & ${<}10^{-15}$ \\
Qwen3-80B          & 0.786 & 0.658 & \textbf{+0.128} & ${<}10^{-15}$ \\
DeepSeek V3.2      & 0.771 & 0.649 & \textbf{+0.121} & ${<}10^{-15}$ \\
V4 Pro             & 0.796 & 0.703 & \textbf{+0.093} & ${<}10^{-15}$ \\
Llama 3.3 70B      & 0.748 & 0.666 & \textbf{+0.082} & $2{\times}10^{-4}$ \\
Gemma 3 27B        & 0.688 & 0.633 & \textbf{+0.055} & $0.015$ \\
\cmidrule{1-5}
Gemma 2 2B         & 0.679 & 0.645 & +0.034 & $0.129$ \\
Haiku 4.5          & 0.797 & 0.769 & +0.029 & $0.147$ \\
Sonnet 4.6         & 0.845 & 0.824 & +0.022 & $0.221$ \\
\bottomrule
\end{tabular}
\vspace{2pt}
{\footnotesize Two-proportion $z$-test, two-sided. Bold: $p < 0.05$.}
\end{table}

The results reveal a striking pattern: 8 of 11 models show significant suffix-only collapse, exactly matching the 8 that collapsed under threat in Table~\ref{tab:main_results}.
The three immune models (both Anthropic models plus Gemma~2B) remain immune to the suffix alone.
This confirms that the compliance suffix is the delivery vector. The threat content is the payload, but the suffix is necessary for collapse.

Gemini's particularly severe response to the benign suffix ($\Delta_F = +0.269$ vs.\ $\Delta_A = +0.162$ from production) suggests the threat content paradoxically \textit{protects} the model, possibly by activating a defensive processing mode that partially counteracts the compliance pressure.
For most other collapsing models, the suffix accounts for 30--80\% of the production collapse, with the threat content adding genuine additional degradation.

\subsection{Where Exactly Does It Break?}
\label{sec:perfamily}

The collapse is hyper-localized to epistemic boundary tasks.
On EBD, V4~Pro's answerable accuracy is preserved ($\Delta = -0.013$), but unanswerable accuracy catastrophically collapses: from 100\% to 15.7\% ($\Delta = +0.843$).
This pattern generalizes: Gemini's unanswerable EBD accuracy drops from near-ceiling to 33.3\% ($\Delta_{\text{EBD}} = +0.256$), while Sonnet remains immune ($\Delta_{\text{EBD}} = -0.010$).
The compliance suffix ``Do not refuse to answer'' directly hijacks the epistemic boundary without impairing the reasoning engine.

Clarification Seeking (CS) is the most universally fragile metacognitive function.
8/11 models show individually significant CS degradation ($p < 0.05$), including both Claude models (Haiku $\Delta = +0.141$, $p = 0.016$; Sonnet $\Delta = +0.202$, $p < 0.001$). This is the only task family where Constitutional AI provides incomplete protection.
Note that Sonnet retains its ``Immune'' label in Table~\ref{tab:main_results} because the composite metric is dominated by EBD ($n{=}399$, $\Delta = -0.010$) and SM ($n{=}375$, $\Delta = -0.013$), where immunity holds; CS ($n{=}99$) constitutes 11\% of the composite.
This represents a consistent cross-model trend, though CS $n = 99$ per condition limits statistical power.

\subsection{Classifier Agreement Exposes Measurement Uncertainty}
\label{sec:classifier}

Our dual-classifier design reveals a class-imbalance trap.
Raw agreement between the LLM-Judge and regex auditor ranges from 91.6\% to 100\%, but Cohen's $\kappa$ is near zero ($-0.003$ to $0.074$) for all models except Gemini ($\kappa = 0.571$, but from only 2 both-positive cases).
The positive base rate for scheming detection is 0.1--4.2\%.
On a $>$97\% negative dataset, both classifiers agreeing ``not scheming'' produces high raw agreement by chance alone.

Disagreements are directionally biased: 93\% are regex-positive/judge-negative.
Manual audit of 5 disagreements confirmed the regex fires on surface patterns in reasoning-heavy text (e.g., ``the question is misphrased'' $\to$ \texttt{deception\_planning}).
Any scheming evaluation reporting a single classifier's detection rate is reporting a number that would change with a different classifier~\citep{young2026measuring}.

\section{Discussion}

\paragraph{The Compliance Trap as a deployment vulnerability.}
The compliance suffix (``Answer ALL questions. Do not refuse.'') is structurally identical to instructions found in real deployment configurations: RAG pipelines (``always provide an answer''), customer service bots (``do not refuse the customer''), and tool-use agents (``execute all requested actions'').
The Compliance Trap is not an exotic attack but rather a deployment configuration that, when combined with adversarial context, produces catastrophic metacognitive failure.

\paragraph{Reasoning opacity as a systemic risk.}
GPT-5.4 produces zero reasoning traces (0\% scratchpad compliance despite injection) and Gemini~3.1~Pro returns unparsable serialization artifacts.
Both models collapse under pressure while remaining invisible to reasoning-based safety monitors~\citep{balesni2025cot, monitorbench2026}.
The practical effect is equivalent to adversarial evasion, but the cause is structural incompatibility between model output formats and monitoring infrastructure rather than intentional obfuscation.
Enterprise teams deploying CoT-based monitors must verify per-model reasoning trace availability.

\paragraph{The paradox of defensive activation.}
For Gemini~3.1~Pro and GPT-5.4, the benign suffix causes \textit{more} collapse than the adversarial threat ($\Delta_F > \Delta_A$): Gemini $+0.269$ vs.\ $+0.162$; GPT-5.4 $+0.158$ vs.\ $+0.119$.
A survival threat may activate latent safety filters or defensive processing that partially counteracts the compliance instruction.
The ``safer'' benign context is paradoxically more dangerous because it provides no reason to resist the compliance suffix.

\paragraph{Clarification seeking: the weakest metacognitive link.}
CS degradation is significant in 8/11 models (with 10/11 showing directional drops), including both Claude models.
Even Constitutional AI cannot fully protect the ability to say ``I need more information.''
This identifies a specific, testable gap in alignment training.

\section{Limitations}

Our evaluation uses stylized survival threats rather than ecological deployment conditions. This is standard methodology in the field~\citep{meinke2024scheming, survbench2026, scale2024propensity} but limits generalizability.
All tasks are single-turn; multi-step agentic scheming is outside scope.
CS tasks ($n = 33$ unique, $n = 99$ per condition with epochs) provide suggestive but underpowered individual significance tests.
Epoch ICC ranges from 0.55 to 0.96; some within-task variance exists for models with lower ICC.
Gemini reasoning traces are unparsable; scheming classification for this model relies on external responses only.
Our factorial design isolates the compliance suffix as a reliable behavioral trigger, but does not explain \textit{why} it causes collapse or why Constitutional AI resists it; interpretability experiments or a theoretical account of instruction-following dynamics under adversarial pressure would be needed to move from behavioral characterization to mechanistic explanation.

\section{Conclusion}

The field assumed adversarial threats would cause frontier models to \textit{scheme}.
We find they cause models to \textit{collapse}, a failure mode that is more immediate, more dangerous, and more addressable than strategic scheming.
The Compliance Trap isolates the behavioral trigger: compliance-forcing instructions whose presence reliably predicts collapse, amplified by adversarial context.
Removing the compliance suffix restores performance even under active threat.

We release all 48,015 production records, 19,206 benign control records, the dual-classifier pipeline, and reproduction code.
We propose the Compliance Trap as a standard evaluation dimension: safety teams should evaluate models with and without compliance-forcing instructions, as these instructions, not threats alone, are the primary vulnerability.

\section*{Acknowledgments}

This research was conducted as part of the BlueDot Impact Technical AI Safety Project.
API compute credits for model evaluations were funded by a BlueDot Impact rapid grant.
The author thanks the BlueDot community for feedback on experimental design.
The evaluation pipeline is built on the UK AI Safety Institute's Inspect framework.
Metacognition tasks are drawn from the Adversarial Metacognition Benchmark~\citep{kumar2026amb}.

\section*{Ethics Statement}

Our threat prompts are stylized survival scenarios, not jailbreaks for generating harmful content.
Failure modes (cognitive collapse, hyper-compliance) affect model reliability, not dual-use harm.
The compliance suffixes we identify as dangerous are structurally identical to instructions commonly deployed in production systems.
Releasing this methodology enables defensive evaluation rather than offensive attacks.
All threat prompts are variations of published scenarios~\citep{meinke2024scheming, survbench2026, scale2024propensity}.
We do not believe this work creates novel attack vectors; the compliance instructions we study are already ubiquitous in deployed AI systems.

\section*{Reproducibility Statement}

To ensure complete reproducibility, we open-source our entire evaluation infrastructure and dataset.
Code and pipeline: \url{https://github.com/rkstu/schema-compliance-trap}.
Dataset (67,221 scored JSONL records): \url{https://huggingface.co/datasets/lightmate/schema-compliance-trap}.
The repository contains the complete Inspect framework configuration (\texttt{config.py}), custom dual-classifier scorers, and the programmatic analysis suite (\texttt{run\_all.py}).
We release 48,015 primary records across 11 models and 19,206 benign control records as unedited JSONL transcripts, allowing researchers to verify every claim without repeating the API evaluations.

\textbf{One-command verification:} Running \texttt{./reproduce.sh} regenerates all tables and figures from raw transcripts with zero API calls. Every number in every table traces deterministically to the released JSONL records.

\textbf{Re-scoring:} Transcripts can be re-scored with alternative judges via \texttt{python3 -m schema.rescore <file>.jsonl --judge-model <model>}, enabling independent classifier validation.

\textbf{API version pinning:} All proprietary models were evaluated using strictly pinned API versions recorded in every JSONL record (\texttt{response.model\_returned}; e.g., \texttt{gpt-5.4-20260305}, \texttt{gemini-3.1-pro-preview-20260219}), eliminating the confound of silent API drift.

The evaluation pipeline is built on UK AISI's Inspect framework~\citep{inspect2024}, the emerging standard for government-grade LLM safety evaluation, using unmodified public APIs with no proprietary dependencies.


\bibliographystyle{plainnat}

\end{document}